\documentclass[11pt]{article}

\usepackage[margin=1in]{geometry}
\usepackage[authoryear,round]{natbib}
\usepackage{amsmath}
\usepackage{amssymb}
\usepackage{enumitem}
\usepackage{amsthm}
\usepackage{bm}
\usepackage{algorithm}      
\usepackage{algpseudocode}   % 提供伪代码的排版语法（新版，推荐）
\usepackage{newtxtext}   % Times 风格字体（IEEE 用的就是 Times/Nimbus Roman）
\usepackage{microtype}
\allowdisplaybreaks
\usepackage[colorlinks=true, linkcolor=blue, citecolor=blue]{hyperref}

\newtheorem{lemma}{Lemma}
\newtheorem{theorem}{Theorem}
\newtheorem{proposition}{Proposition}
\newtheorem{remark}{Remark}
\newtheorem{definition}{Definition}
\def\cN{{\cal N}}
\title{A Theoretical Perspective on DDPM: The Denoising Process as an Exact Reverse-Time Diffusion }
\author{%
Yuanqiu Mo\textsuperscript{1},
Wangzhi Zhou\textsuperscript{1},
Soura Dasgupta\textsuperscript{2}\\[6pt]
{\small \textsuperscript{1}School of Mathematics, Southeast University,
Nanjing 211189, China}\\
{\small \texttt{yuanqiumo@seu.edu.cn}, \texttt{wangzhizhou474@gmail.com}}\\[4pt]
{\small \textsuperscript{2}Department of Electrical and Computer Engineering,
University of Iowa, Iowa City, IA 52242, USA}\\
{\small \texttt{soura-dasgupta@uiowa.edu}}
}
\date{}

\begin{document}

\maketitle
\begin{abstract}
Diffusion models have achieved remarkable success in generative models, yet the theoretical foundations of their discrete-time denoising processes remain less fully understood. The score-based SDE framework \cite{song2019generative} unifies DDPM as a discretization of a continuous-time reverse-time SDE, and exactness in that framework is a property of the continuous-time limit rather than of the discrete sampler. In this paper we study a complementary notion: exactness with respect to the discrete forward process, i.e., whether the discrete-time forward diffusion process admits a bijective reverse built from an independent backward noise sequence \citep{dasgupta2026reversetimediffusionprocessesdiscrete}. We first show that the denoising process of DDPM, when the true initial data $x_0$ is substituted, coincides with the exact discrete-time reverse of the DDPM forward diffusion process. We then develop a generalized framework for discrete-time forward diffusion processes and derive the corresponding exact reverse mapping and sampling-based reverse procedure. Furthermore, we show that the proposed framework encompasses the closed-form reverse processes of both DDPM and score-based models with Langevin dynamics (SMLD). These results provide a unified discrete-time perspective on exact reverse diffusion and offer a systematic theoretical foundation for constructing reverse processes in diffusion models.
\end{abstract}
\section*{Introduction}
Diffusion models have recently emerged as a powerful class of generative models, demonstrating remarkable success in capturing complex and high-dimensional data distributions \citep{ho2020denoising}. The forward diffusion process iteratively adds Gaussian noise to the sampled data, whereas the reverse (denoising) process is employed to reconstruct the original data. By training a neural network to predict the injected noise (or equivalently, the score function of the data distribution)\footnote{Denoising, score matching, and the variational
lower bound are equivalent views of the same training objective. See
\citet{luo2022understanding} for a unified treatment.}, diffusion models can generate high-fidelity samples from pure Gaussian noise. Subsequent work learned the reverse variances and adopted a cosine noise schedule,
improving likelihood and sampling efficiency~\citep{nichol2021improved}. Owing to this exceptional expressiveness, diffusion models have been widely applied to various domains. Representative examples include new benchmarks for high-quality images synthesis in computer vision \citep{dhariwal2021diffusion, rombach2022high}, highly expressive policy classes for capturing complex, multi-modal action distributions in offline reinforcement learning (RL) \citep{wang2023diffusion, zhang2024entropy}, and powerful sampling-based solvers that leverage known system dynamics for trajectory optimization \citep{pan2024mbd}.

The reverse process in DDPM aims to reconstruct the original data by iteratively sampling from the conditional distribution \(p_\theta(x_{t-1}\,|\, x_t)\) using the reparameterization trick \citep{kingma2013auto}. The conditional distribution is parameterized by a neural network trained to maximize the variational lower bound (ELBO) \citep{ho2020denoising}. While this mechanism has been empirically successful, its theoretical foundation is elegantly unified by the score-based SDE framework \citep{song2021score}, which reveals that the DDPM reverse process is intrinsically connected to the continuous-time reverse-time stochastic differential equation (SDE) originally derived by \citet{anderson1982reverse}. %However, it is important to note that this equivalence is only approximate. 
Specifically, the discrete-time forward diffusion process is first converted into a continuous-time SDE by taking a continuous-time limit. A reverse-time SDE is then derived using the theory developed in \cite{anderson1982reverse}, and discretizing this reverse-time SDE yields the corresponding reverse difference equation. The continuous-time reverse SDE is exact only in the limit of infinitesimal time steps, whereas the practical reverse process in DDPM is implemented in discrete time through numerical discretization. This transition from continuous to discrete time inevitably introduces approximation errors (e.g., due to Taylor truncation), thereby compromising the exactness of reverse sampling and potentially altering the resulting data distribution. More fundamentally, properties established for continuous-time SDEs do not necessarily carry over to their discrete-time counterparts. As rigorously demonstrated in recent work \citep{dasgupta2026reversetimediffusionprocessesdiscrete}, it has long been established \citep{clements1973well} that certain SDEs with bounded and stable solutions can exhibit unbounded or even explosive behavior when subjected to standard numerical discretization. Consequently, the continuous-time SDE framework \citep{song2021score} provides a unified theoretical perspective, while its practical discrete-time implementations may require additional analysis to characterize the effects of discretization.

Despite these practical advantages and intuitive interpretations, a rigorous discrete-time characterization of the conditions under which the DDPM denoising step constitutes an exact reverse process, as well as a systematic approach to constructing such a reverse process, has remained largely implicit.
Inspired by \citet{dasgupta2026reversetimediffusionprocessesdiscrete}, this paper develops a systematic approach to constructing the denoising process from the perspective of exact reverse-time diffusion, which refers to the exact recovery of the original forward diffusion process through a bijective mapping involving independent backward noise sequences that reconstruct the forward trajectory. Building on this notion, we establish that the denoising process of \citet{ho2020denoising} is, in fact, an exact reverse-time diffusion process when the initial sample \(x_0\) is known. Furthermore, we generalize the forward process in \citet{ho2020denoising} and derive the corresponding general form of exact reverse diffusion, together with a sampling-based reverse procedure analogous to that in \citet{ho2020denoising}. We further show that the proposed framework encompasses the closed-form reverse processes of prominent score-based diffusion models, including DDPM \citep{ho2020denoising} and SMLD \citep{song2019generative, song2021score}.

The rest of the paper is organized as follows. Section \ref{sec:posterior} presents the preliminaries of the discrete-time forward diffusion process considered in DDPM and derives the corresponding conditional posterior distribution of the injected noise. Section \ref{sec:exact_reverse_conds} introduces the conditions for exact reverse diffusion, while Section \ref{sec:generate_erp} constructs the reverse process from the perspective of exact reverse-time diffusion. Sections \ref{sec:rddpm} and \ref{sec:exactinter} establish the equivalence between the resulting reverse process and the sampling-based reverse procedure adopted in DDPM. Finally, Section \ref{sec:general} develops a generalized forward diffusion framework and derives its corresponding sampling-based reverse process using the notion of exact reverse-time diffusion.

\section{Posterior Distribution of $u_t$ Given $x_0$ and $x_t$} \label{sec:posterior}

Consider the standard discrete-time forward diffusion process as follows
\begin{equation} \label{eq:diffusion}
    x_t = \sqrt{\alpha_t}x_{t-1} + \sqrt{\beta_t}u_t, \quad \forall\, 1 \leq t \leq T,
\end{equation}
where $x_t \in \mathbb{R}^n, u_t \in \mathbb{R}^n$, \(u_t \sim \mathcal{N}(0, I)\) is the independent forward noise and $\alpha_t = 1 - \beta_t$. The noise scales are prescribed such that $x_T$ is approximately distributed according to $\mathcal{N}(0, I)$. Moreover, from \eqref{eq:diffusion}, we derive that 
\begin{align}
    x_t &= \sqrt{\alpha_t}(\sqrt{\alpha_{t-1}}x_{t-2}+\sqrt{\beta_{t-1}}u_{t-1}) + \sqrt{\beta_t}u_t \notag \\
    &=\sqrt{\alpha_t \alpha_{t-1}}x_{t-2}+ \sqrt{\alpha_t \beta_{t-1}}u_{t-1}+\sqrt{\beta_t}u_t \notag\\
    &=\cdots \notag\\
    &= \sqrt{\bar \alpha_t}x_0+ \sum_{i=1}^t \sqrt{(\prod_{j=i+1}^t\alpha_j) \beta_i } \cdot u_i, \label{eq:induction_x_0}
\end{align}
where \eqref{eq:induction_x_0} uses induction with \(\bar \alpha_t = \prod_{i=1}^t \alpha_i\). Thus, the marginal distribution given  \(x_0\) is
\begin{equation}\label{eq:marginal_distribution}
    x_t \mid x_0 \sim \mathcal{N}\left(\sqrt{\bar{\alpha}_t}x_0, (1-\bar{\alpha}_t)I\right),
\end{equation}
where \eqref{eq:marginal_distribution} trivially holds from \(\mathbb{E}[x_t | x_0] = \sqrt{\bar{\alpha}_t}x_0\) and \(\text{Var}(x_t | x_0) = (1-\bar{\alpha}_t)I\). To find the posterior distribution of \(u_t\) given \(x_t\) and \(x_0\), we first establish the joint distribution of \((u_t, x_t)\) conditioned on \(x_0\). By the definition of \(u_t\) and \eqref{eq:marginal_distribution}, the cross-covariance is derived as:
\begin{align}
    \text{Cov}(u_t, x_t \mid x_0) &= \text{Cov}\left(u_t, \sqrt{\alpha_t}x_{t-1} + \sqrt{\beta_t}u_t \,\middle |\, x_0\right) \notag\\
    &= \sqrt{\alpha_t}\text{Cov}(u_t, x_{t-1} \mid x_0) + \sqrt{\beta_t}\text{Cov}(u_t, u_t \mid x_0) \notag\\
    &=\sqrt{\beta_t}
I,\label{eq:u_t_given_x_0_covariance}
\end{align}
where \eqref{eq:u_t_given_x_0_covariance} follows from the fact that \(u_t\) is independent of the past state \(x_{t-1}\) and \(x_0\), i.e.,      \(\text{Cov}(u_t, x_{t-1} | x_0) = 0\). Therefore, the joint distribution is a multivariate Gaussian shown as follows:
\begin{equation} \label{eq:multi-gaussian}
    \begin{pmatrix} u_t \\ x_t \end{pmatrix} \Bigg|\, x_0 \sim \mathcal{N} \left( \begin{pmatrix} \mathbf{0} \\ \sqrt{\bar{\alpha}_t}x_0 \end{pmatrix}, \begin{pmatrix} I & \sqrt{\beta_t}I \\ \sqrt{\beta_t}I & (1-\bar{\alpha}_t)I \end{pmatrix} \right)
\end{equation}
\eqref{eq:multi-gaussian} further reveals that the conditional distribution of $U_t$ given $X_0$ and $X_t$ is multivariate Gaussian with a diagonal covariance matrix.
\begin{lemma}\label{le:posterior}
    Given $\tilde{X} = (X_t, X_0)$, then 
    \begin{equation}
    p_{U_t|\tilde{X}}(u_t|\tilde{x}) = \prod_{i=1}^n \mathcal{N}\left(u_t^{(i)} \mid \mu_{u_t|\tilde{x}}^{(i)}, \sigma_t^2\right),
    \end{equation}
     where 
    \begin{equation} \label{eq:conditional_mean_variance}
        \sigma_t^2 = \frac{\alpha_t(1-\bar{\alpha}_{t-1})}{1-\bar{\alpha}_t} ,~ \mu_{u_t|\tilde{x}} = \frac{\sqrt{\beta_t}}{1-\bar{\alpha}_t}(x_t - \sqrt{\bar{\alpha}_t}x_0),
    \end{equation} 
$\bar{\alpha}_0 := 1$, and $u_t^{(i)}$ denotes the $i$-th component of $u_t$.
\end{lemma}
\begin{proof}
The proof is provided in Appendix \ref{sec:lemma1}.
\end{proof}

\section{Exact Reverse Conditions} \label{sec:exact_reverse_conds}
In this section, we derive conditions under which the forward diffusion process (\ref{eq:diffusion}) admits an exact reverse. To this end, we first extend the notion of an exact reverse introduced in \citet{dasgupta2026reversetimediffusionprocessesdiscrete} to the setting where $x_0$ is given.
\newpage
\begin{definition}\label{def:condition}
    Consider the forward diffusion process (\ref{eq:diffusion}) and the following reverse process:
    \begin{equation}\label{eq:reverse}
        x_{t-1} = R_{t}(x_{t}, x_0, v_{t}), 
        \quad \forall\, 2 \leq t \leq T,
    \end{equation}
    where the backward noise sequence is generated according to
    \begin{equation}\label{eq:renoise}
        v_t = H_t(x_t, x_0, u_{t}), 
        \quad \forall\, 2 \leq t \leq T,
    \end{equation}
    with $R_{t}: \mathbb{R}^{n}\times\mathbb{R}^{n}\times\mathbb{R}^{m}\to\mathbb{R}^{n}$ and
    $H_t: \mathbb{R}^{n}\times\mathbb{R}^{n}\times\mathbb{R}^{m}\to\mathbb{R}^{m}$ continuously differentiable function in both arguments.
    We say that (\ref{eq:reverse}) and (\ref{eq:renoise}) constitute an exact reverse of (\ref{eq:diffusion}) given $x_0$ if the following conditions are satisfied:
    \begin{enumerate}
        \item The forward diffusion (\ref{eq:diffusion}) is recovered by (\ref{eq:reverse}) under (\ref{eq:renoise}), and the mapping between $(x_{t-1},u_t)$ and $(x_{t},v_{t})$ is bijective.
        \item $v_t$ is independent of $x_j$ for all $j\geq t$.
        \item The random variables $v_2,\ldots,v_T$ are jointly independent.
    \end{enumerate}
    Note that no backward noise is introduced at $t=1$, since $u_1$ is uniquely determined by $(x_1,x_0)$ through (\ref{eq:diffusion}), making the terminal reverse step deterministic. %(see subsequently Remark~\ref{re:vanish}).
\end{definition}
%To establish that the reverse process is exact, merely satisfying the independence condition in Theorem 3 is not sufficient. We must also ensure that the mapping between the forward and reverse states is a bijection. This requires combining our derivations with Lemma 2 and Theorem 1 from \citet{dasgupta2026reversetimediffusionprocessesdiscrete}.
Likewise, the sufficient conditions for establishing the existence of an exact reverse process for \eqref{eq:diffusion} given $x_0$ can be extended from those in \citet[Theorem 1]{dasgupta2026reversetimediffusionprocessesdiscrete}.

%We first revisit Theorem 1 of \cite{dasgupta2026reversetimediffusionprocessesdiscrete}, which establishes sufficient conditions for the existence of an exact reverse process for \eqref{eq:diffusion}.  %exists if the following two conditions are satisfied:
\begin{theorem}\label{thm:conditions}
The forward diffusion process in \eqref{eq:diffusion} admits an exact reverse process, as defined in Definition \ref{def:condition}, if the following two conditions are satisfied:
\begin{enumerate}[label=\Roman*.]
    \item (Parametrized bijection condition) There
exists a continuously differentiable mapping $\psi_{t}: \mathbb{R}^{n}\times\mathbb{R}^{n}\times\mathbb{R}^{m}\to\mathbb{R}^{m}$ such that $u_t = \psi_{t}(x_{t}, x_0, v_{t})$ holds\footnote{In \citet[Theorem 1]{dasgupta2026reversetimediffusionprocessesdiscrete}, the reverse process is conditioned on $x_t$ alone, whereas in our setting, it is conditioned on $\tilde{x}=(x_t,x_0)$. Consequently, at $t=1$, $u_1$ is uniquely determined by $(x_1,x_0)$. Thus, the construction only needs to consider $t\geq 2$.};
\item (State and reverse-time noise independence condition) For each $t$, $v_t$ generated by (\ref{eq:renoise}) and $x_t$ are independent random variables.
\end{enumerate}
\end{theorem}
By exploiting the component-wise conditional independence of the components of $U_t$ given $\tilde{X}$, as established in Lemma \ref{le:posterior}, we construct the Rosenblatt transformation $V_t=h(\tilde{X},U_t)$ component-wise using the corresponding conditional CDFs \citep{rosenblatt1952remarks}. This construction ensures that both Conditions I and II of Theorem \ref{thm:conditions} are satisfied.
\begin{proposition}\label{prop:conditions}
    Consider the forward diffusion process (\ref{eq:diffusion}). For $t \ge 2$, let the backward noise random vector be $V_t = h(\tilde{X},U_t)$, whose $i$-th component is given by
\begin{equation}\label{eq:h}
    h_i(\tilde{x}, u_t^{(i)}) = P(U_t^{(i)} \le u_t^{(i)} | \tilde{X} = \tilde{x}) = \Phi\left( \frac{u_t^{(i)} - \mu_{u_t|\tilde{x}}^{(i)}}{\sigma_t} \right),
\end{equation}
where $\sigma_t$ and $\mu_{u_t|\tilde x}$ are given in (\ref{eq:conditional_mean_variance}). Then (\ref{eq:diffusion}) admits an exact reverse process.
\end{proposition}
\begin{proof}
    The proof is given in the Appendix \ref{sec:prop1}.
\end{proof}
%\begin{remark}\label{re:vanish}
%    The construction is degenerate at $t=1$: $\sigma_1^2=0$ because $\bar\alpha_0=1$. %so $u_1$ is determined exactly by $(x_1,x_0)$ and the reverse step $x_0=\frac{1}{\sqrt{\alpha_1}}(x_1-\sqrt{\beta_1}u_1)$ is deterministic. 
%    We will show later that this is consistent with the standard DDPM convention of setting $z=0$ at $t=1$ \citep[Algorithm 2]{ho2020denoising}. %and with the $\sigma_0=0$ degeneracy in the SMLD case. We therefore state all results for $t\ge 2$ and handle $t=1$ separately.
%\end{remark}

%. Combined with Lemma 1, this guarantees that \(V_t\) is a valid independent white noise process. Therefore, the reverse process is exact.

\section{Generating Samples By Exact Reverse Process Construction} \label{sec:generate_erp}
Equations \eqref{eq:h} and \eqref{eq:one} imply that $V_t$ is uniformly distributed over $[0,1]^n$, i.e., $V_t\sim\mathcal{U}([0,1]^n)$ for $t \ge 2$. This result enables the exact reversal of the forward diffusion process in \eqref{eq:diffusion} through sampling. Since Gaussian noise is more commonly used in sampling procedures and offers analytical convenience, we transform \(V_t\) into a standard Gaussian random vector via the inverse standard normal CDF, \(\Phi^{-1}(\cdot)\) \citep{dasgupta2026reversetimediffusionprocessesdiscrete,rosenblatt1952remarks}. Since this transformation is a componentwise bijective deterministic mapping, it preserves the independence of the components. Accordingly, defining the reverse noise as \(Z_t = \Phi^{-1}(V_t)\), we obtain:
\begin{equation}\label{eq:zgaussian}
    Z_t \sim \mathcal{N}(0, I), \quad \forall\, t \geq 2
\end{equation}
and \(Z_t\) remains independent of \(\tilde{X} = (X_t, X_0)\). Hereafter, we use lowercase \(z_t \in \mathbb{R}^n\) to denote a particular realization of the random vector \(Z_t\) for practical implementation. By directly inverting the CDF in \eqref{eq:h}, we can recover the exact noise realization from the corresponding transformed uniform random variable, namely,
\begin{equation} \label{eq:noise}
    u_t = \mu_{u_t|\tilde{x}} + \sigma_t z_t = \frac{\sqrt{\beta_t}}{1-\bar{\alpha}_t}(x_t - \sqrt{\bar{\alpha}_t}x_0) + \sigma_t z_t, 
\end{equation}
where the conditional mean $\mu_{u_t|\tilde{x}}$ is given by \eqref{eq:conditional_mean_variance}. In particular, when $t = 1$, $u_1$ reduces to $\frac{\sqrt{\beta_1}}{1-\bar{\alpha}_1}(x_1 - \sqrt{\bar{\alpha}_t}x_0)$ since $\sigma_1 = 0$, i.e., $u_1$ is determined exactly by $(x_1,x_0)$. Substituting the resulting exact realization of $u_t$ into the forward diffusion equation
$x_{t-1} = \frac{1}{\sqrt{\alpha_t}}
    (x_t - \sqrt{\beta_t}u_t)$
establishes a bijective mapping between $z_t$ and $x_{t-1}$, with $x_t$ and $x_0$ given. Consequently, for $t \geq 2$, the reverse sampling equation for $x_{t-1}$ is given by
\begin{align}
    x_{t-1} &= \frac{1}{\sqrt{\alpha_t}}\left(x_t-\sqrt{\beta_t} \cdot (\mu_{u_t|\tilde{x}} + \sigma_t z_t)\right) \notag\\
    &= \underbrace{\frac{1}{\sqrt{\alpha_t}}(x_t-\sqrt{\beta_t} \cdot \mu_{u_t|\tilde{x}})}_{\tilde \mu(x_t, x_0)} - \underbrace{\frac{\sqrt{\beta_t}\sigma_t}{\sqrt{\alpha_t}}}_{\tilde \sigma_t}z_t = \tilde \mu(x_t, x_0) - \tilde \sigma_t z_t,
    \label{eq:exact_reverse_function1}
\end{align}
whereas for $t=1$, $x_0$ is exactly recovered as $\tilde{\mu}(x_1,x_0)$.
Thus, based on \eqref{eq:exact_reverse_function1}, we can readily construct an exact reverse sampling function given by
$x_{t-1} = \tilde{\mu}_t(x_t,x_0) - \tilde{\sigma}_t z_t,$
provided that the initial data $x_0$ is completely known. Note that, since $Z_t \sim \mathcal{N}(0,I)$, the negated noise $-Z_t$ has the same distribution as $Z_t$. Therefore, we can equivalently rewrite \eqref{eq:exact_reverse_function1} as
\begin{equation} \label{eq:exact_reverse_function}
        x_{t-1} = \tilde{\mu}_t(x_t,x_0) + \tilde{\sigma}_t z_t, \quad \forall\, t \geq 2
\end{equation}
which is the form adopted in the remainder of this paper.

\section{Reverse Process in Denoising Diffusion Probabilistic Model}\label{sec:rddpm}
The classical DDPM framework \citep{ho2020denoising} adopts the same forward diffusion process as defined in \eqref{eq:diffusion} and constructs the reverse denoising process according to
\begin{equation}\label{eq:ddpm}
    x_{t-1} = \frac{1}{\sqrt{\alpha_t}}
    \left(x_t - \frac{1-\alpha_t}{\sqrt{1-\bar{\alpha}_t}}
    \epsilon_{\theta}(x_t,t)\right) + \tilde{\sigma}_t \tilde z_t, \quad\forall\, t \geq 1,
\end{equation}
where $\tilde{\sigma}_t$ is given in (\ref{eq:exact_reverse_function}), $\tilde Z_t \in \cN(0, I)$ for $t \geq 2$, $\tilde z_1 := 0$, and $\epsilon_{\theta}(x_t,t)$ predicts the noise from the current noisy sample $x_t$ and the diffusion step $t$. Theoretically, the training objective is to maximize the evidence lower bound (ELBO). In practice, however, \citet{ho2020denoising} introduce a simplified mean squared error (MSE) objective, which avoids assigning excessively small weights to early diffusion steps, thereby improving their contribution to training, as follows:
\begin{equation} \label{eq:train_loss}
    L_{\text{simple}}(\theta) := \mathbb{E}_{t, x_0, \epsilon} \left[ \left\| \epsilon - \epsilon_\theta\left(\sqrt{\bar{\alpha}_t}x_0 + \sqrt{1-\bar{\alpha}_t}\epsilon, t\right) \right\|^2 \right],
\end{equation}
where $\epsilon_\theta(x_t,t)$ is a neural network parameterized by $\theta$ that takes the noisy sample $x_t$ and diffusion step $t$ as inputs and estimates the noise $\epsilon$ introduced by the forward process. Trained using the above MSE objective, this network serves as the core noise-prediction component of the reverse denoising process. %A similar design is adopted by \citet{zhang2024entropy} to estimate the initial data or the clean action in reinforcement learning. 
In what follows, we show that minimizing the noise prediction MSE also minimizes the MSE of the resulting estimate of the initial data $x_0$.

%A critical distinction must be made between the \textit{inference-time algebraic manipulation} and the \textit{training objective} of diffusion models. 
Consider the standard simplified noise-matching MSE loss in \eqref{eq:train_loss}.
%While \cite{zhang2024entropy} explicitly introduces an estimate of the clean action \(\hat{x}_0\) during the reverse process, its underlying neural network \(\epsilon_\theta\) shares an identical training paradigm with the standard DDPM \citep{ho2020denoising}. Both networks are optimized via the simplified noise-matching MSE loss denoted in \eqref{eq:train_loss}
At first glance, this objective suggests that the neural network is trained solely to predict the noise introduced during the forward diffusion process. However, applying the reparameterization trick \citep{kingma2013auto} to (\ref{eq:marginal_distribution}) yields
\(x_t = \sqrt{\bar{\alpha}_t}x_0 + \sqrt{1-\bar{\alpha}_t}\epsilon\), where \(\epsilon \sim \mathcal{N}(0,I)\). Substituting the network output \(\epsilon_\theta\) for \(\epsilon\), we obtain the estimated \(\hat{x}_0\)
%can be predicted directly through the predicted noise $\epsilon_{\theta}$
as
\begin{equation} \label{eq:hat_x_0}
    \hat x_0 = \frac{x_t - \sqrt{1 - \bar \alpha_t} \epsilon_{\theta}(x_t,t)}{\sqrt{\bar \alpha_t}}.
\end{equation}
Thus, the true noise and the predicted noise can be rewritten in terms of $x_0$ and $\hat x_0$ respectively as
\begin{equation}
    \epsilon = \frac{x_t - \sqrt{\bar{\alpha}_t}x_0}{\sqrt{1-\bar{\alpha}_t}}, \quad \text{and} \quad \epsilon_\theta = \frac{x_t - \sqrt{\bar{\alpha}_t}\hat{x}_0}{\sqrt{1-\bar{\alpha}_t}}.
    \label{eq:noise_to_x0}
\end{equation}
Note that the estimate $\hat{x}_0=(x_t-\sqrt{1-\bar{\alpha}_t}\epsilon_\theta)/\sqrt{\bar{\alpha}_t}$ in \eqref{eq:hat_x_0} is equivalent to the discrete-time counterpart of (8) in \citet{zhang2024entropy}. The underlying neural network $\epsilon_\theta$ in their formulation is trained using the same paradigm as in the standard DDPM. Substituting \eqref{eq:noise_to_x0} into \eqref{eq:train_loss} yields:
\begin{align}
    L_{\text{simple}}(\theta) = \mathbb{E}_{t, x_0, \epsilon} \left[ \left\| \frac{x_t - \sqrt{\bar{\alpha}_t}x_0}{\sqrt{1-\bar{\alpha}_t}} - \frac{x_t - \sqrt{\bar{\alpha}_t}\hat{x}_0}{\sqrt{1-\bar{\alpha}_t}} \right\|^2 \right]= \mathbb{E}_{t, x_0, \epsilon} \left[ \frac{\bar{\alpha}_t}{1-\bar{\alpha}_t} \left\| \hat{x}_0 - x_0 \right\|^2 \right]. \label{eq:loss_equiv}
\end{align}
As shown in \eqref{eq:loss_equiv}, for a given time step \(t\), the coefficient \(\frac{\bar{\alpha}_t}{1-\bar{\alpha}_t}\) is a constant. Therefore, minimizing the noise prediction error is equivalent to minimizing the clean sample prediction error.

\section{Exact Reverse Diffusion in DDPM and Beyond}\label{sec:beyond}
In this section, we show that the reverse sampling process in DDPM, which is constructed based on the reparameterization technique \citep{kingma2013auto}, is in fact an exact reverse-time diffusion process associated with the forward process \eqref{eq:diffusion}, provided that the initial data $x_0$ is completely known. Meanwhile, we provide a theoretical justification for the choice of the noise variance parameter $\tilde{\sigma}_t^2 = \frac{1-\bar{\alpha}_{t-1}}{1-\bar{\alpha}_t}\beta_t,$ adopted in the reverse process of DDPM (see Algorithm 2 in \citet{ho2020denoising}). Finally, we extend the exact reverse sampling construction to more general settings, including cases in which $x_{t-1}$ and $u_t$ have different dimensions.
\subsection{An Exact Reverse-Time Diffusion Interpretation of DDPM}\label{sec:exactinter}
Consider the proposed reverse process in \eqref{eq:exact_reverse_function}. By substituting $\hat{x}_0=(x_t-\sqrt{1-\bar{\alpha}_t}\epsilon_\theta)/\sqrt{\bar{\alpha}_t}$ into $\tilde{\mu}_t(x_t,x_0)$ in \eqref{eq:exact_reverse_function}, where the neural network $\epsilon_\theta$ is used to estimate the forward noise as in DDPM, we obtain that for $t \geq 2$,
\begin{align}
    x_{t-1} 
    &= \tilde{\mu}_t(x_t, \hat{x}_0) + \tilde{\sigma}_t z_t = \frac{\sqrt{\alpha_t}(1-\bar{\alpha}_{t-1})}{1-\bar{\alpha}_t}x_t + \frac{\sqrt{\bar{\alpha}_{t-1}}\beta_t}{1-\bar{\alpha}_t}\hat{x}_0 + \tilde{\sigma}_t z_t \notag\\
    &= \frac{\sqrt{\alpha_t}(1-\bar{\alpha}_{t-1})}{1-\bar{\alpha}_t}x_t + \frac{\sqrt{\bar{\alpha}_{t-1}}\beta_t}{1-\bar{\alpha}_t} \left[ \frac{1}{\sqrt{\bar{\alpha}_t}}\left(x_t - \sqrt{1-\bar{\alpha}_t}\epsilon_\theta\right) \right] + \tilde{\sigma}_t z_t \notag\\
    &= \frac{\sqrt{\alpha_t}(1-\bar{\alpha}_{t-1})}{1-\bar{\alpha}_t}x_t + \frac{\sqrt{\bar{\alpha}_{t-1}}\beta_t}{\sqrt{\bar{\alpha}_t}(1-\bar{\alpha}_t)}x_t - \frac{\sqrt{\bar{\alpha}_{t-1}}\beta_t\sqrt{1-\bar{\alpha}_t}}{\sqrt{\bar{\alpha}_t}(1-\bar{\alpha}_t)}\epsilon_\theta + \tilde{\sigma}_t z_t \notag\\
   &= \frac{\sqrt{\alpha_t}(1-\bar{\alpha}_{t-1})}{1-\bar{\alpha}_t}x_t + \frac{\beta_t}{\sqrt{\alpha_t}(1-\bar{\alpha}_t)}x_t - \frac{\beta_t}{\sqrt{\alpha_t}\sqrt{1-\bar{\alpha}_t}}\epsilon_\theta + \tilde{\sigma}_t z_t \notag\\
    &= \left[ \frac{\sqrt{\alpha_t}(1-\bar{\alpha}_{t-1})}{1-\bar{\alpha}_t} + \frac{\beta_t}{\sqrt{\alpha_t}(1-\bar{\alpha}_t)} \right] x_t - \frac{\beta_t}{\sqrt{\alpha_t}\sqrt{1-\bar{\alpha}_t}}\epsilon_\theta + \tilde{\sigma}_t z_t \notag\\
    &= \left[ \frac{\alpha_t(1-\bar{\alpha}_{t-1})}{\sqrt{\alpha_t}(1-\bar{\alpha}_t)} + \frac{\beta_t}{\sqrt{\alpha_t}(1-\bar{\alpha}_t)} \right] x_t - \frac{\beta_t}{\sqrt{\alpha_t}\sqrt{1-\bar{\alpha}_t}}\epsilon_\theta + \tilde{\sigma}_t z_t \notag\\
    &= \frac{\alpha_t(1-\bar{\alpha}_{t-1}) + \beta_t}{\sqrt{\alpha_t}(1-\bar{\alpha}_t)} x_t - \frac{\beta_t}{\sqrt{\alpha_t}\sqrt{1-\bar{\alpha}_t}}\epsilon_\theta + \tilde{\sigma}_t z_t \notag\\
    &= \frac{1-\bar{\alpha}_t}{\sqrt{\alpha_t}(1-\bar{\alpha}_t)} x_t - \frac{\beta_t}{\sqrt{\alpha_t}\sqrt{1-\bar{\alpha}_t}}\epsilon_\theta + \tilde{\sigma}_t z_t \notag\\
    &= \frac{1}{\sqrt{\alpha_t}} x_t - \frac{\beta_t}{\sqrt{\alpha_t}\sqrt{1-\bar{\alpha}_t}}\epsilon_\theta + \tilde{\sigma}_t z_t \notag\\
    &= \frac{1}{\sqrt{\alpha_t}} \left( x_t - \frac{\beta_t}{\sqrt{1-\bar{\alpha}_t}}\epsilon_\theta \right) + \tilde{\sigma}_t z_t. \label{eq:equivalence_reverse}
\end{align}
where the derivation makes use of the identities $\bar{\alpha}_t=\alpha_t\bar{\alpha}_{t-1}$, $\beta_t=1-\alpha_t$, and $\alpha_t(1-\bar{\alpha}_{t-1})+\beta_t=1-\bar{\alpha}_t$. For $t = 1$, by substituting $\hat{x}_0=(x_1-\sqrt{1-\bar{\alpha}_1}\epsilon_\theta)/\sqrt{\bar{\alpha}_1}$ into $\tilde{\mu}_1(x_1,x_0)$, it holds that
\begin{equation}\label{eq:t1case}
    x_0 = \frac{1}{\sqrt{\alpha_1}} \left( x_1 - \frac{\beta_1}{\sqrt{1-\bar{\alpha}_1}}\epsilon_\theta \right).
\end{equation}
Since \eqref{eq:equivalence_reverse} and \eqref{eq:t1case} are identical to \eqref{eq:ddpm}, this establishes the equivalence between the proposed reverse process and Algorithm 2 of DDPM. Note that DDPM assumes Gaussian reverse transitions, i.e.,
$p_\theta(x_{t-1}\,|\,x_t)=\mathcal N(\mu_\theta(x_t,t),\Sigma_\theta(x_t,t))$.
However, the true reverse transition distribution $p(x_{t-1}\,|\,x_t)$ is a mixture over $x_0$ and is therefore not Gaussian in general. In contrast, the conditional distribution $p(x_{t-1}\,|\,x_t,x_0)$ is exactly Gaussian. Moreover, although setting the variance in DDPM as \(\tilde{\sigma}_t^2 = \frac{1-\bar{\alpha}_{t-1}}{1-\bar{\alpha}_t}\beta_t\) is optimal for a deterministically fixed \(x_0\) \citep{ho2020denoising}, it also yields an exact reverse process when \(x_0\) is known.

%Our method according to \eqref{eq:exact_reverse_function} gives
%\begin{equation}\label{eq:variance_equivalence}
%    \tilde{\sigma_t}^2 = \frac{\beta_t}{\alpha_t} \sigma_t^2 = \frac{\beta_t}{\alpha_t} \cdot \frac{\alpha_t(1-\bar{\alpha}_{t-1})}{1-\bar{\alpha}_t} = \frac{1 - \bar \alpha_{t-1}}{1 - \bar \alpha_{t}} \beta_t = \tilde \beta_t.
%\end{equation}

\begin{remark}
    (\ref{eq:equivalence_reverse}) and (\ref{eq:ddpm}) together establish the equivalence between our exact reverse diffusion process in (\ref{eq:exact_reverse_function}) and the denoising process of \citet{ho2020denoising}, despite their fundamentally different objectives. Specifically, the former is derived based on the notion of an exact reverse process given in Definition~\ref{def:condition}, whereas the latter is designed to maximize ELBO. %revealing the intrinsic construction procedure of the denoising process, and how the mean and the variance come.
    In particular, while the backward noise $\tilde z_1$ in (\ref{eq:ddpm}) is set to zero in DDPM \citep[Algorithm 2]{ho2020denoising}, from the perspective of exact reverse, $x_0$ is recovered exactly as $\tilde{\mu}_1(x_1,x_0)$ as in (\ref{eq:exact_reverse_function}). Accordingly, in our case, $x_0$ is obtained by evaluating $\tilde \mu_1(x_1,x_0)$ with $x_0$ replaced by its estimate $\hat{x}_0$, without introducing any backward noise.
\end{remark}
\subsection{More General Cases}\label{sec:general}
We now consider a generalized version of \eqref{eq:diffusion}, in which the forward diffusion process is given by
\begin{equation}\label{eq:general-diffusion}
    x_t = Ax_{t-1} + Bu_t, \quad \forall\, 1 \leq t \leq T,
\end{equation}
where $A \in \mathbb{R}^{n\times n}$ and $B \in \mathbb{R}^{n\times n}$ are both of full column rank. Note that the full column rank condition on \(B\) is necessary for \(u_t\) to be uniquely determined by \(x_t\) and \(x_{t-1}\) through an implicit function that is continuously differentiable in both arguments. %Moreover, in this generalized setting, $x_{t-1}$ and $u_t$ are allowed to have different dimensions. 

To construct a valid reverse process, it is crucial to specify the distribution of the terminal state $x_T$, which serves as the starting point for the reverse process. As the forward process progresses, the repeated application of the transition matrix $A$ and the injection of independent noise $Bu_t$ accumulate a total covariance $\Sigma_T$. Thus, the terminal state follows a Gaussian distribution:
%Assuming the spectral radius of $A$ is strictly less than 1, i.e., the process \eqref{eq:general-diffusion} is stable, the terminal state converges to a Gaussian distribution:
\begin{equation} \label{eq:terminal_state_distribution}
    x_T \sim \mathcal{N}(0,\Sigma_T),
\end{equation}
where $\Sigma_T = \sum\nolimits_{k=0}^{T-1} A^k B B^\top (A^\top)^k$ is derived in \eqref{eq:cale}. Specifically, for DDPM, it can be readily verified that \(\Sigma_T\) converges to the identity matrix \(I\) as \(T\to\infty\). %For SMLD where $A = I$, it accumulates to $\sigma_T^2I$.

\begin{table}[t]
\vspace{-1em}
\centering
\renewcommand{\arraystretch}{1.2} % 稍微拉开一点行距
\algrenewcommand\algorithmicindent{0.5em}

\begin{tabular}{@{} p{0.40\textwidth} @{\hspace{2em}} p{0.55\textwidth} @{}}
% 左侧算法框（较窄）
\begin{minipage}{0.42\textwidth}
\small
\begin{algorithm}[H]
\caption{Training}
\label{alg:general_training}
\begin{algorithmic}[1]
\Repeat
    \State \(x_0 \sim q(x_0)\)
    \State \(t \sim \text{Uniform}(\{1, \dots, T\})\)
    \State \(\epsilon \sim \mathcal{N}(\mathbf{0}, I)\)
    \State Take gradient descent step on
    \State \(\nabla_\theta \|\epsilon - \epsilon_\theta(\bar{A}_t x_0 + \Sigma_t^{1/2} \epsilon, t)\|^2\)
\Until{converged}
\vspace{0.11em} 
\end{algorithmic}
\end{algorithm}
\end{minipage} 
& 
% 右侧算法框（较宽）
\begin{minipage}{0.55\textwidth}
\small
\begin{algorithm}[H]
\caption{Sampling}
\label{alg:general_sampling}
\begin{algorithmic}[1]
\State \(x_T \sim \mathcal{N}(\mathbf{0}, \Sigma_T)\)
\For{\(t = T, \dots, 2\)}
    \State \(z \sim \mathcal{N}(\mathbf{0}, I)\) 
    \State \(L_t = \text{Cholesky}(I - B_t^\top \Sigma_t^{-1} B_t)\)
    \State \(x_{t-1} = A^{-1}_t ( x_t - B_t ( B_t^\top \Sigma_t^{-1/2} \epsilon_\theta(x_t, t) + L_t z ))\)
\EndFor
\vspace{-0.15em} 
\State \Return \(A_1^{-1}(x_1-\Sigma_1^{1/2} \epsilon_{\theta}(x_1,1))\)
\end{algorithmic}
\end{algorithm}
\end{minipage} \\
\end{tabular}
\end{table}
\begin{theorem}\label{thm:general_exact_reverse}
    Consider the general forward process (\ref{eq:general-diffusion}) and the reverse noise construction of $v_t \sim \mathcal{U}([0,1]^n)$ specified in Section \ref{sec:generate_erp}. Define  $\Sigma_t := \sum\nolimits_{k=0}^{t-1} A^k B B^\top (A^\top)^k$ and $\Sigma_{u_t|\tilde x} = I - B^\top \Sigma_t^{-1}B$, and suppose that $\Sigma_{u_t|\tilde{x}}$ is positive definite for $t \geq 2$. Then there exists a continuously differentiable bijective mapping $\psi_t$ such that $u_t = \psi_t(\tilde x, v_t)$, and the exact reverse of (\ref{eq:general-diffusion}) is given by
    \begin{equation} \label{eq:general_exact_reverse}
        x_{t-1} = R_t(\tilde x, v_t) = A^{-1}(x_t - B \psi_t(\tilde x,v_t)), \quad \forall\, 2\le t\le T,
    \end{equation}
    where $\psi_t(\tilde{x},v_t)$ admits the following explicit expression:
    \begin{equation} \label{eq:general_psi_t}
        \psi_t(\tilde x, v_t) = B^\top \Sigma_t^{-1} (x_t - \bar A_t x_0) + L_t \Phi^{-1}(v_t),
    \end{equation}
    with $\bar{A}_t:=A^t$, $L_t$ denoting the lower-triangular Cholesky factor of $\Sigma_{u_t|\tilde{x}}$, i.e., $\Sigma_{u_t|\tilde{x}}=L_tL_t^\top$ for all $t \geq 2$, and $\Phi^{-1}(\cdot)$ denoting the inverse standard normal cumulative distribution function. %The case $t=1$ is deterministic and involves no such factorization (see Remark~\ref{re:time-varying}).
\end{theorem}
\begin{proof}The proof is given in the Appendix \ref{sec:psi_construction}.
\end{proof}
%It's also noteworthy that at $t=1$ the reverse step is deterministic. Since $x_1=Ax_0+Bu_1$ with $B$ of full column rank, $u_1$ is uniquely determined by $(x_1,x_0)$, and the step reduces to $x_0=A^{-1}(x_1-Bu_1)$. Accordingly, \eqref{eq:general_exact_reverse}--\eqref{eq:general_psi_t} are applied for $2\le t\le T$ only: the last step involves neither a random draw nor a Cholesky factorization of $\Sigma_{u_1\mid\tilde x}$ (cf.~Remark~\ref{re:vanish}).
\begin{remark}\label{re:time-varying}
    To ensure that \(\Sigma_{u_t|\tilde{x}}\) in Theorem \ref{thm:general_exact_reverse} is positive definite for all \(t\geq 2\), %(note that the conditional covariance vanishes identically at \(t=1\); see similarly Remark \ref{re:vanish}), 
    one can choose \(A\) and \(B\) to be diagonal matrices, with all diagonal entries of \(A\) belonging to \((0,1]\) and those of \(B\) being nonzero.
    This choice is consistent with the standard setups of DDPM and SMLD \citep{song2019generative}, which will be introduced later. 
    
    Furthermore, Theorem \ref{thm:general_exact_reverse} can also be extended to a time-varying case. Consider a forward process with time-varying, full-column-rank matrices $A_t \in \mathbb{R}^{n\times n} $ and $B_t \in \mathbb{R}^{n\times n}$:
    \begin{equation} \label{eq:time-varying_forward_diffusion}
        x_t = A_tx_{t-1} + B_tu_t, \quad \forall\, 1 \leq t \leq T,
    \end{equation}
    then Theorem \ref{thm:general_exact_reverse} holds similarly with the same closed form in (\ref{eq:general_exact_reverse}) and (\ref{eq:general_psi_t}) by substituting $A_t$, $B_t$ and $L_t$ into the corresponding items with $\bar A_t = \prod_{i=1}^t A_i$ and $ \Sigma_t = \sum_{k=1}^t \left( \prod_{i=k+1}^t A_i \right) B_k B_k^\top \left( \prod_{i=k+1}^t A_i \right)^\top.$
\end{remark}
Since \(\Phi^{-1}(v_t)\) follows a standard Gaussian distribution, by using the forward noise MSE estimator \(\epsilon_\theta\) to estimate the initial data based on the reparameterization of (\ref{eq:cale}), we obtain the point estimate 
\begin{equation}\label{eq:general-data-estimate}
    \hat x_0 = \bar A_t^{-1}(x_t - \Sigma_t^{1/2} \epsilon_{\theta}(x_t,t))
\end{equation}
Substituting this into the posterior mean $\mu_{u_t|\tilde x} = B^\top\Sigma_t^{-1}(x_t-\bar A_t x_0)$ derived in \eqref{eq:meva} yields the sampling-based reverse procedure presented in Algorithm \ref{alg:general_sampling}. In particular, $x_0$ is recovered by applying \eqref{eq:general-data-estimate} to $x_1$ obtained from the reverse process, without introducing backward noise, since $x_0$ is uniquely determined by $x_1 = A_1x_0 + \Sigma_1^{1/2}\epsilon$ with $\epsilon \sim \cN(0, I)$ by the reparameterization of \eqref{eq:cale}. The training procedure for \(\epsilon_\theta\), which is essentially the same as that in DDPM, is provided in Algorithm \ref{alg:general_training}.

\begin{remark}
    The conditioning on $x_0$ can be recast as an exact reverse of a Markov chain by carrying the initial state as a frozen component, which is the classical device behind fixed-point smoothing \citep{moore1971fixed}. Define $\tilde X_t = \tilde x = (x_t,x_0) \in \mathbb{R}^{2n}$, $\tilde A_t = \text{diag}(A_t,I)$ and $\tilde B_t = (B_t^\top,0)^\top$ with $A_t$ and $B_t$ shown in (\ref{eq:time-varying_forward_diffusion}). The proposed time-varying forward process (\ref{eq:time-varying_forward_diffusion}) can be rewritten as
    \begin{equation}
        \tilde X_{t} = \tilde A_t \tilde X_{t-1} + \tilde B_t u_{t}, \quad \forall \,1 \le t \le T.
    \end{equation}
   Then, the sequence ${\tilde X_t}$ forms a Markov chain driven by the same noise sequence ${u_t}$ as in (\ref{eq:time-varying_forward_diffusion}). At each reverse step, only the $x_t$ component of $\tilde{x}$ is updated, while the remaining $x_0$ component is kept unchanged. Based on Theorem \ref{thm:conditions} and Remark \ref{re:time-varying}, the exact reverse process can be constructed recursively by setting $v_t = \Phi\left(L_t^{-1}(u_t-\mu_{u_t|\tilde x})\right)$ for $2\le t\le T$
   where $L_t$ and $\mu_{u_t|\tilde x}$ are given in Remark \ref{re:time-varying} and (\ref{eq:meva}) (with the coefficients updated to the time-varying case), respectively. At $t=2$, the resulting $\tilde{X}_1$ already yields $x_0$. This construction assumes that $x_0$ is known. When $x_0$ is unknown, the same recursion can be implemented using Algorithm \ref{alg:general_sampling}, with $x_0$ estimated via $\epsilon_{\theta}$.
\end{remark}

To further demonstrate the generality of our framework, we extend the discussion of \eqref{eq:general-diffusion} and Theorem \ref{thm:general_exact_reverse} to Score-based Models with Langevin Dynamics (SMLD) \citep{song2019generative}. Similar to DDPM, SMLD employs a forward perturbation process characterized by a predefined sequence of monotonically increasing positive noise scales, \(0 < \sigma_1 < \sigma_2 < \dots < \sigma_T\). By sequentially perturbing the initial data point \(x_0\) according to these noise scales, we obtain
\begin{equation}\label{eq:SMLD_forward}
    x_t = x_{t-1} + \sqrt{\sigma_t^2 - \sigma_{t-1}^2}u_t, \quad\, \forall\,1\le t\le T,
\end{equation}
where \(\sigma_0 = 0\), \(x_t \in \mathbb{R}^n\), and \(u_t \in \mathbb{R}^n\) is an independent standard Gaussian noise vector, i.e., \(u_t \sim \mathcal{N}(0,I)\). Iterating \eqref{eq:SMLD_forward} yields
\begin{equation} \label{eq:SMLD_x_t_given_x_0_distribution}
    x_t \mid x_0 \sim \mathcal{N}(x_0,\sigma_t^2 I).
\end{equation}
Unlike DDPM, where $x_T\sim\mathcal{N}(0,I)$, the terminal sample $x_T$ in SMLD follows $\mathcal{N}(0,\sigma_T^2 I)$, with $\sigma_T$ being a predefined noise scale. As in \eqref{eq:SMLD_forward} and \eqref{eq:SMLD_x_t_given_x_0_distribution}, SMLD takes $A = I$, $B = \sqrt{\sigma_t^2 - \sigma_{t-1}^2} I$, $\bar A_t = I$ and $\Sigma_t = \sigma_t^2I$ to match the coefficients in \eqref{eq:general-diffusion}. By the general reverse form from Theorem \ref{thm:general_exact_reverse} in \eqref{eq:general_exact_reverse} and \eqref{eq:general_psi_t}, we obtain
\begin{equation} \label{eq:smld_reverse_psi}
    x_{t-1} = x_t - \sqrt{\sigma_t^2 - \sigma_{t-1}^2} \cdot \psi_t(\tilde x,v_t),
\end{equation}
where $\psi_t(\tilde x, v_t)$ is specified by 
\begin{equation} \label{eq:psi_t_appro}
    \psi_t(\tilde x, v_t) = \frac{\sqrt{\sigma_t^2 - \sigma_{t-1}^2}}{\sigma_t} \epsilon + \frac{\sqrt{\tilde \beta_t}}{\sqrt{\sigma_t^2 - \sigma_{t-1}^2}} \Phi^{-1}(v_t),
\end{equation}
with $\tilde \beta_t = \sigma^2_{t-1}(\sigma_t^2-\sigma_{t-1}^2)/\sigma_t^2$ and $\Phi^{-1}(\cdot)$ being the inverse normal CDF mapping. The detailed construction of \eqref{eq:psi_t_appro} is provided in Appendix \ref{sec:smldpsi_construction}. As $\Phi^{-1}(v_t) \sim \cN(0, I)$, the sign of the noise term can be flipped as in \eqref{eq:exact_reverse_function1}--\eqref{eq:exact_reverse_function}, and it follows from (\ref{eq:ddpm}), \eqref{eq:smld_reverse_psi} and \eqref{eq:psi_t_appro} that SMLD admits a DDPM-like reverse sampling process given as follows, which has also been proposed in \citet{song2021score} 
\begin{equation}\label{eq:SMLD-reverse}
    x_{t-1} = x_t - \frac{\sigma_t^2 - \sigma_{t-1}^2}{\sigma_t}\epsilon_\theta(x_t,t) + \sqrt{\frac{\sigma_{t-1}^2(\sigma_t^2 - \sigma_{t-1}^2)}{\sigma_t^2}}z_t,\quad \, \forall\,1\le t\le T,
\end{equation}
where \(\epsilon_\theta(x_t,t)\) can be chosen as the same noise-prediction network adopted in DDPM and $z_t \sim \cN(0, I)$ is the sampled Gaussian noise.

Notably, given \(x_0\), this reverse process is also \textit{exact} with respect to the forward process in \eqref{eq:SMLD_forward}. %since the coefficients in \eqref{eq:induction_x_0} can be analytically mapped to the form in \eqref{eq:SMLD_forward}. 
\textit{Therefore, the proposed generalized framework for exact reverse diffusion can also be directly applied to the SMLD formulation.}

\section*{Conclusion}
%This paper provides an exact reverse-time diffusion process corresponding to the forward diffusion \eqref{eq:diffusion}. We demonstrate that this reverse-time process is mathematically equivalent to the denoising process in DDPM \cite{ho2020denoising}, thereby closing the theoretical loop: \textit{Given \(x_0\), the denoising process in DDPM is theoretically rigorous from the perspective of an exact reverse process---a connection that has not been previously explained or verified.} Note that a key step in our reverse construction relies on the independence between \(V_t\) and \(\tilde{X}\), which is guaranteed by designing the transformation function \(h\) in (\ref{eq:h}) as the CDF of \(u_t|\tilde{x}\). In fact, such a transformation function is not unique, and future work could explore alternative constructions of \(h\). Moreover, since the exact reverse relies on the assumption that the initial data \(x_0\) is known, estimating the distribution of \(x_0\) could be another direction for future work.

This paper first provides an exact reverse-time diffusion process for the forward diffusion (\ref{eq:diffusion}), and shows that the DDPM denoising process coincides with it whenever the initial state $x_0$ is known. A key ingredient is the independence of the transformed backward noise $V_t=h(\tilde X,U_t)$, or equivalently $Z_t=\Phi^{-1}(V_t)$, which is obtained by choosing $h$ as the conditional CDF of $u_t|\tilde x$, implying that  $V_t$ and $X_j$ are independent for all $j\ge t$ and that $V_2,\dots,V_T$ are mutually independent (the terminal step $t=1$ carries no backward noise). It's noteworthy that the transformation $h$ is not unique, and alternative constructions are left for future work. We further derive the exact reverse for general linear–Gaussian forward diffusion process, which covers DDPM and the SMLD as special cases. Accordingly, since the reverse variance is induced by $A$ and $B$, refinements in \citet{nichol2021improved} may prove more effective here. Moreover, the established reverse construction is exact when the initial data \(x_0\) is known. In practice, however, \(x_0\) is replaced by its network-based estimate \(\hat{x}_0\). Thus, characterizing how the estimation error \(\hat{x}_0-x_0\) affects the resulting reverse process and the associated sampling distribution is therefore another natural direction for future research.

%the correlation structure of the noise collapses into a scaled identity matrix under \(\tilde{X} = (X_t, X_0)\), making the components independent, which analytically fulfills Theorem 3 and guarantees an exact reverse via Theorem 1 and 4.
\bibliographystyle{chicago}
\bibliography{ref}

\appendix
\section{Proof of Lemma \ref{le:posterior}}\label{sec:lemma1}
Using the standard formula for conditional Gaussian distributions (Bayes' rule for Gaussians), where \(y|x \sim \mathcal{N}(\mu_y + \Sigma_{yx}\Sigma_x^{-1}(x-\mu_x), \Sigma_y - \Sigma_{yx}\Sigma_x^{-1}\Sigma_{xy})\). 
To compute the posterior distribution of $u_t$, we identify the corresponding variables conditioned on $x_0$ by setting $y=u_t$ and $x=x_t$ in the above expression. By \eqref{eq:multi-gaussian}, we derive that \(\mu_{y} = \mathbb{E}[u_t|x_0] = \mathbf{0}\) and \(\mu_{x} = \mathbb{E}[x_t|x_0] = \sqrt{\bar \alpha_t} x_0\), while \(\Sigma_y = \text{Var}(u_t|x_0) = I\) and \(\Sigma_x = \text{Var}(x_t|x_0) = (1-\bar \alpha_t)I\) with \(\Sigma_{yx} = \Sigma_{xy}^\top = \text{Cov}(u_t,x_t|x_0) = \sqrt{\beta_t}I\). Thus, simply set \(\tilde x = (x_t, x_0)\), it holds that the posterior distribution of $u_t$ given both $x_t$ and $x_0$ is exactly \(\mathcal{N}(\mu_{u_t|\tilde{x}}, \Sigma_{u_t|\tilde{x}})\), the conditional mean and the conditional covariance of which are specified as 
\begin{equation}\label{eq:mean}
    \mu_{u_t|\tilde{x}} = \mathbf{0} + \sqrt{\beta_t}I \left[ (1-\bar{\alpha}_t)I \right]^{-1} \left( x_t - \sqrt{\bar{\alpha}_t}x_0 \right) = \frac{\sqrt{\beta_t}}{1-\bar{\alpha}_t}(x_t - \sqrt{\bar{\alpha}_t}x_0)
\end{equation}
and
\begin{align}
    \Sigma_{u_t|\tilde{x}} &= I - \sqrt{\beta_t}I \left[ (1-\bar{\alpha}_t)I \right]^{-1} \sqrt{\beta_t}I = \left( 1 - \frac{\beta_t}{1-\bar{\alpha}_t} \right) I \notag\\
    &= \frac{1-\bar{\alpha}_t-\beta_t}{1-\bar{\alpha}_t} I = \frac{\alpha_t(1-\bar{\alpha}_{t-1})}{1-\bar{\alpha}_t}I,\label{eq:substitute}
\end{align}
respectively, where \eqref{eq:substitute} comes from the identity \(1-\bar{\alpha}_t = \alpha_t(1-\bar{\alpha}_{t-1}) + \beta_t\). We further express the posterior covariance as:
\begin{equation}
    \Sigma_{u_t|\tilde{x}} = \sigma_t^2 I, \label{eq:covariance}
\end{equation}
where $\sigma_t^2 = \frac{\alpha_t(1-\bar{\alpha}_{t-1})}{1-\bar{\alpha}_t}$.
It can be observed from \eqref{eq:covariance} that the components of the noise vector \(U_t\) are mutually independent given \( \tilde{X} = \tilde{x} = (x_t, x_0)\). Consequently, the joint posterior density admits the following factorization:
\begin{equation*}
    p_{U_t|\tilde{X}}(u_t|\tilde{x}) = \prod_{i=1}^n \mathcal{N}\left(u_t^{(i)} \mid \mu_{u_t|\tilde{x}}^{(i)}, \sigma_t^2\right).
\end{equation*}

\section{Proof of Proposition \ref{prop:conditions}}\label{sec:prop1}
We first provide the function $\psi_{t}$ required by Condition I. 
For any fixed \(\tilde{x}\), since the conditional distribution of \(U_t\) is Gaussian with positive variance \(\sigma_t^2 > 0\), each \(h_i(\tilde{x}, \cdot)\) is strictly monotonic and continuous. Thus, the inverse mapping \(h^{-1}(\tilde{x}, \cdot)\) exists everywhere. This ensures the existence of a continuously differentiable function \(\psi_t\) such that:
\begin{equation}
    U_t = \psi_t(\tilde{X}, V_t),
\end{equation}
and Condition I is satisfied.

We next verify Condition II. Define the Jacobian of $h$ in (\ref{eq:h}) with respect to $u$ as $J_{hu}: \mathbb{R}^{n} \times \mathbb{R}^{n} \to \mathbb{R}^{n} \times \mathbb{R}^{n}$, the $kl$-th element of which obeys
\begin{equation}\label{eq:Jacobian}
    [J_{hu}(\tilde{x}, u_t)]_{kl} = \frac{\partial h_k(\tilde{x}, u_t)}{\partial u^{(l)}_t}.
\end{equation}
Since \(h_i\) depends exclusively on \(u_t^{(i)}\), the Jacobian matrix \(J_{hu}\) is purely diagonal. Its determinant is exactly the product of the diagonal entries such that
\begin{equation}
    \det [J_{hu}(\tilde{x}, u_t)] = \prod_{i=1}^n \frac{\partial h_i}{\partial u_t^{(i)}} = \prod_{i=1}^n p_{U_t^{(i)}|\tilde{X}}(u_t^{(i)} | \tilde{x}) = p_{U_t|\tilde{X}}(u_t|\tilde{x}) . \label{eq:det}
\end{equation}
Under a mild assumption that $p_{U_t, \tilde{X}}(u_t, \tilde{x}) > 0$ everywhere, we have $\det [J_{hu}(\tilde{x}, u_t)] \neq 0$ for all $\tilde{x}$ and $u_t$. Then, by the change-of-variables formula for conditional densities, we obtain
\begin{equation}\label{eq:one}
    p_{V_t|\tilde{X}}(h(\tilde{x}, u_t)|\tilde{x}) = \frac{p_{U_t|\tilde{X}}(u_t|\tilde{x})}{|\det [J_{hu}(\tilde{x}, u_t)]|} = 1, 
\end{equation}
where the last equality follows from (\ref{eq:det}). Therefore, the conditional density of $V_t$ is actually  invariant to the conditioning variable $\tilde{X} = (X_t, X_0)$, which establishes independence of $V_t$ and $X_t$.
%Substituting \eqref{eq:det} into Theorem 3, the fraction becomes exactly \(1\). Thus, we can set \(\xi(v) = 1\), satisfying the necessary and sufficient condition. This proves that \(V = h(\tilde{X}, U_t)\) is independent of \(\tilde{X}\). Furthermore, Theorem 4 strictly guarantees that \(V\) is uniformly distributed on \([0,1]^m\).

Since both Condition I and Condition II are satisfied, Theorem \ref{thm:conditions} guarantees the exact reversal of the forward diffusion process, with the backward noise sequence generated according to (\ref{eq:h}).

\section{Proof of Theorem \ref{thm:general_exact_reverse}}\label{sec:psi_construction}
%We will first show the form as in \eqref{eq:general_exact_reverse}, and then provide a detailed derivation of the general exact mapping function \(\psi_t(\tilde{x}, v_t)\) in \eqref{eq:general_psi_t}, as well as the specific form in \eqref{eq:psi_t_appro}.
By recursively expanding the general forward process in \eqref{eq:general-diffusion}, we can express \(x_t\) in terms of \(x_0\) and the noise variables \(u_1,\ldots,u_t\) as
\begin{align}
    x_t &= Ax_{t-1}+Bu_t \notag\\
    &= A(Ax_{t-2}+Bu_{t-1})+Bu_t \notag\\
    &=\cdots \notag\\
    &=\bar A_t x_0+\sum_{\ell=0}^{t-1}A^{t-\ell-1}Bu_{\ell+1},
    \label{eq:general_x_t_expansion}
\end{align}
where \(\bar A_t\) is defined in \eqref{eq:general_psi_t}. Accordingly, the conditional covariance of \(x_t\) given \(x_0\) is given by
\begin{align}
    \Sigma_t
    &=\operatorname{Var}(x_t\mid x_0) \notag\\
    &=\operatorname{Var}\left(
    \bar A_t x_0+
    \sum_{\ell=0}^{t-1}A^{t-\ell-1}Bu_{\ell+1}
    \,\middle|\,x_0\right) \notag\\
    &=\sum_{\ell=0}^{t-1}A^\ell BB^\top(A^\ell)^\top \label{eq:cale}. 
\end{align}
Moreover, since \(u_t\) is independent of \(x_i\) for all \(i<t\), we have
\begin{align}
    \operatorname{Cov}(u_t,x_t\mid x_0)
    &=\operatorname{Cov}(u_t,Ax_{t-1}+Bu_t\mid x_0) \notag\\
    &=\operatorname{Var}(u_t)B^\top
    =B^\top.
\end{align}
Similarly, $\operatorname{Cov}(x_t,u_t\mid x_0)=B.$ 
Furthermore, it follows directly from \eqref{eq:general_x_t_expansion} that
$\mathbb E[x_t\mid x_0]=\bar A_t x_0.$
Consequently, the joint conditional distribution of \((u_t,x_t)\) given \(x_0\) is Gaussian and can be expressed as
\begin{equation}
    \begin{pmatrix} u_t \\ x_t \end{pmatrix}
    \Bigg|\, x_0
    \sim
    \mathcal N\left(
    \begin{pmatrix}
        \mathbf{0}\\
        \bar A_t x_0
    \end{pmatrix},
    \begin{pmatrix}
        I & B^\top\\
        B & \Sigma_t
    \end{pmatrix}
    \right).
\end{equation}
By the conditional Gaussian formula, the posterior distribution of \(u_t\) given \(\tilde x=(x_t,x_0)\) is also Gaussian, namely, $u_t\mid\tilde x
\sim
\mathcal N\left(
\mu_{u_t\mid\tilde x},
\Sigma_{u_t\mid\tilde x}
\right),$
where
\begin{equation}\label{eq:meva}
    \mu_{u_t\mid\tilde x}
    =B^\top\Sigma_t^{-1}(x_t-\bar A_t x_0),
    \qquad
    \Sigma_{u_t\mid\tilde x}
    =I-B^\top\Sigma_t^{-1}B.
\end{equation}
%Following the derivation of the conditional Gaussian posterior in Section \ref{sec:generate_erp}, the joint distribution of $(x_t, u_t)$ conditioned on $x_0$ is Gaussian. 
For general matrices $A$ and $B$, since $\Sigma_{u_t|\tilde{x}}$ is symmetric positive definite, it admits a Cholesky decomposition $\Sigma_{u_t|\tilde{x}}=L_tL_t^\top,$
where $L_t$ is a lower-triangular matrix. Define the whitened variable $\xi_t = L_t^{-1}(u_t - \mu_{u_t|\tilde x})$, we trivially have $\mathbb{E}[\xi_t|\tilde x] = L_t^{-1}(\mu_{u_t|\tilde x}-\mu_{u_t|\tilde x}) = \mathbf{0}$ and $\text{Cov}(\xi_t|\tilde x) = L_t^{-1} \text{Cov}(u_t | \tilde x)L_t = L_t^{-1} (L_tL_t^\top)L_t^{-\top} = I$. Thus, this construction illustrates that $\xi_t|\tilde x \sim \mathcal{N}(\mathbf{0},I)$, implying that $\xi_t$ is independent of $\tilde x$. Furthermore, since $\text{Cov}(\xi_t|\tilde x) =I$ is an identity matirx, the components of $\xi_t$ are mutually independent, with each component satisfying $\xi_t^{(k)}\sim\mathcal{N}(0,1)$.

By applying the Rosenblatt transform, we construct $v_t = h(\tilde x, u_t) =  \Phi(L_t^{-1}(u_t - \mu_{u_t|\tilde x})) = \Phi (\xi_t)$, where $\Phi(\cdot)$ is the normal CDF mapping. By the probability integral transform, each component \(v_t^{(k)} = \Phi(\xi_t^{(k)})\) follows a standard uniform distribution \(\mathcal{U}[0,1]\). Since the components of \(\xi_t\) are independent, the components of \(v_t\) are jointly independent, yielding \(v_t \sim \mathcal{U}([0,1]^n)\). Furthermore, as a deterministic mapping of the \(\tilde{x}\)-independent variable \(\xi_t\), \(v_t\) remains strictly independent of \(\tilde{x}\).

Similar to the analysis in Proposition \ref{prop:conditions}, to verify Condition II of Theorem \ref{thm:conditions}, we compute the Jacobian matrix \(J_{hu}\) of the mapping \(h\) with respect to \(u_t\). Its \((k,l)\)-th entry is given by $[J_{hu}(\tilde{x},u_t)]_{kl}
=\frac{\partial h_k}{\partial u_t^{(l)}}.$
Let \(\phi(\cdot)\) denote the standard normal probability density function (PDF). Applying the chain rule, we obtain
\begin{equation}
    \frac{\partial h_k}{\partial u_t^{(l)}}
    =
    \phi\left((L_t^{-1})_k
    (u_t-\mu_{u_t|\tilde{x}})\right)
    (L_t^{-1})_{kl}.
\end{equation}
where \((L_t^{-1})_k\) and \((L_t^{-1})_{kl}\) denote the \(k\)-th row and the \((k,l)\)-th entry of \(L_t^{-1}\), respectively. Since \(L_t^{-1}\) is lower triangular, we have \((L_t^{-1})_{kl}=0\) for \(l>k\). Consequently, \(J_{hu}\) is lower triangular, and its determinant is given by the product of its diagonal entries:
\begin{equation}
    \det[J_{hu}(\tilde{x}, u_t)] = \prod_{k=1}^n \phi(\xi_t^{(k)}) (L_t^{-1})_{kk}.
\end{equation}
On the other hand, the joint conditional density \(p_{U_t|\tilde{X}}(u_t|\tilde{x})\) can be expressed in terms of the whitened variable \(\xi_t\) as
\begin{align}
    p_{U_t|\tilde{X}}(u_t|\tilde{x}) &= \frac{1}{\sqrt{(2\pi)^n |\det(\Sigma_{u_t|\tilde{x}})|}} \exp\left(-\frac{1}{2} (u_t - \mu_{u_t|\tilde{x}})^\top \Sigma_{u_t|\tilde{x}}^{-1} (u_t - \mu_{u_t|\tilde{x}})\right) \label{eq:decom}\\
    &= \prod_{k=1}^n \phi(\xi_t^{(k)}) \cdot \frac{1}{|\det(L_t)|} = \prod_{k=1}^n \phi(\xi_t^{(k)}) \cdot \prod_{k=1}^n (L_t^{-1})_{kk}.
\end{align}
where \eqref{eq:decom} follows from $\xi_t = L_t^{-1}(u_t - \mu_{u_t|\tilde x})$ and $\Sigma_{u_t|\tilde{x}}=L_tL_t^\top$.
Substituting the above expressions into the change-of-variables formula for conditional densities, we obtain
\begin{equation}
    p_{V_t|\tilde{X}}(v_t|\tilde{x}) = \frac{p_{U_t|\tilde{X}}(u_t|\tilde{x})}{|\det[J_{hu}(\tilde{x}, u_t)]|} = \frac{\prod_{k=1}^n \phi(\xi_t^{(k)}) \prod_{k=1}^n (L_t^{-1})_{kk}}{\prod_{k=1}^n \phi(\xi_t^{(k)}) (L_t^{-1})_{kk}} = 1.
\end{equation}
Hence, the conditional density of \(V_t\) given \(\tilde{X}\) is identically equal to \(1\), independent of \(\tilde{X}\). This establishes that \(V_t\) is independent of \(\tilde{X}\) and verifies Condition II of Theorem \ref{thm:conditions}. Inverting the mapping \(v_t = h(\tilde x, u_t) = \Phi(L_t^{-1}(u_t - \mu_{u_t|\tilde x}))\) yields \(u_t = \psi_t(\tilde{x}, v_t)\). Thus, Condition II of Theorem \ref{thm:conditions} is satisfied, guaranteeing the existence of the exact reverse bijection. Moreover, it follows from (\ref{eq:meva}) that \(\psi_t(\tilde x,v_t)\) admits the following explicit expression:
\begin{equation}
    \psi_t(\tilde x, v_t) = B^\top \Sigma_t^{-1} (x_t - \bar A_t x_0) + L_t \Phi^{-1}(v_t), \label{eq:general_psi_t_validate}
\end{equation}
which establishes \eqref{eq:general_psi_t}.%We then show the specific form of $\psi_t(\tilde x, v_t)$ in \eqref{eq:psi_t_appro}.

\section{Construction of $\psi_t$ in SMLD}\label{sec:smldpsi_construction}
We take the coefficient matrices as $A = I$, $B = \sqrt{\sigma_t^2-\sigma_{t-1}^2}I$, $\bar A_t = I$ and $\Sigma_t = \sigma_t^2I$. It then follows from Theorem \ref{thm:general_exact_reverse} and Remark \ref{re:time-varying} that the conditional covariance is given by \(\Sigma_{u_t|\tilde{x}} = \frac{\sigma_{t-1}^2}{\sigma_t^2}I\), whose Cholesky factor is \(L_t = \frac{\sigma_{t-1}}{\sigma_t}I\). Moreover, since \(\sigma_0 = 0\), iterating the SMLD forward process in \eqref{eq:SMLD_forward} yields $x_t = x_0 + \sigma_t \epsilon$ with $\epsilon \sim \mathcal{N}(0,I).$
Substituting these expressions into \eqref{eq:general_psi_t_validate}, we obtain
\begin{align}
    \psi_t(\tilde x,v_t) &= B^\top (\sigma_t^2I)^{-1}(x_t-x_0) + \frac{\sigma_{t-1}}{\sigma_t} I \cdot\Phi^{-1}(v_t) \notag\\
    &=\frac{\sqrt{\sigma_t^2-\sigma_{t-1}^2}}{\sigma_t^2}(x_t-x_0) + \frac{\sigma_{t-1}}{\sigma_t}\Phi^{-1}(v_t) \notag\\
    &=\frac{\sqrt{\sigma_t^2-\sigma_{t-1}^2}}{\sigma_t}\epsilon + \frac{\sigma_{t-1}}{\sigma_t}\Phi^{-1}(v_t). \label{eq:psi_t_final}
\end{align}
To align the variance term with the standard SMLD posterior variance \(\tilde{\beta}_t\) as in \cite{song2019generative}, we apply the following algebraic identity
\begin{equation} \label{eq:algebraic_identity}
    \frac{\sigma_{t-1}}{\sigma_t} = \frac{\sqrt{\sigma_{t-1}^2(\sigma_t^2 - \sigma_{t-1}^2) / \sigma_t^2}}{\sqrt{\sigma_t^2 - \sigma_{t-1}^2}} = \frac{\sqrt{\tilde{\beta}_t}}{\sqrt{\sigma_t^2 - \sigma_{t-1}^2}},
\end{equation}
where \(\tilde{\beta}_t = \frac{\sigma_{t-1}^2(\sigma_t^2 - \sigma_{t-1}^2)}{\sigma_t^2}\). Substituting \eqref{eq:algebraic_identity} into \eqref{eq:psi_t_final} yields
\begin{equation}
    \psi_t(\tilde{x}, v_t) = \frac{\sqrt{\sigma_t^2 - \sigma_{t-1}^2}}{\sigma_t} \epsilon + \frac{\sqrt{\tilde{\beta}_t}}{\sqrt{\sigma_t^2 - \sigma_{t-1}^2}} \Phi^{-1}(v_t),
\end{equation}
which completes the proof.
\end{document}